# Autonomous Driving System Design for Formula Student Driverless Racecar

hanqing TIAN, Jun NI, Member, IEEE, Jibin HU

*Abstract*—This paper summarizes the work of building the autonomous system including detection system and path tracking controller for a formula student autonomous racecar. A LIDAR-vision cooperating method of detecting traffic cone which is used as track mark is proposed. Detection algorithm of the racecar also implements a precise and high rate localization method which combines the GPS-INS data and LIDAR odometry. Besides, a track map including the location and color information of the cones is built simultaneously. Finally, the system and vehicle performance on a closed loop track is tested. This paper also briefly introduces the Formula Student Autonomous Competition (FSAC) in 2017.

## I. INTRODUCTION

Formula Student China 2017 was held on November 11 in Xiangyang, Hubei Province. The competition is divided into combust engine group, electric group and driverless group which is a new group this year and is named Formula Student Autonomous Competition (FSAC). BIT-FSA team which is one of the first seven domestic teams is invited to participate in the competition. The competition consists of three items: acceleration test which purpose is to test the accelerate performance and detection and control algorithm under high speed; skid pad which test the handling stability; and track drive which is a comprehensive test of perception and control strategy. The *Smart Shark I* autonomous racecar from BIT passed all the technical inspection and accomplished all the three items. Finally, BIT-FSA team won the FSAC champion.

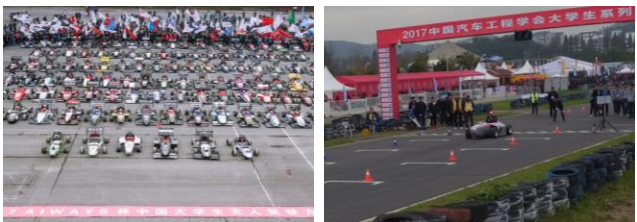

(a) Picture of All Teams      (b) Autonomous Acceleration Test
Fig. 1. Formula Student China 2017

Hanqing Tian is currently a PhD student at National Key Lab of Vehicular Transmission, and School of Mechanical Engineering, Beijing Institute of Technology.
Jun Ni is currently a PhD student at National Key Lab of Vehicular Transmission, and School of Mechanical Engineering, Beijing Institute of Technology. He is working on the dynamics and control of the distributed drive electric vehicles. (nijun_bit@163.com).
Jibin Hu received his PhD from Beijing Institute of Technology in 2002 and is currently a professor at National Key Lab of Vehicular Transmission, and School of Mechanical Engineering, Beijing Institute of Technology.

## II. VEHICLE AND SOFTWARE ARCHITECTURE

The *Smart Shark* autonomous racecar platform was an electric formula racecar (BIT-FSE *Silver Shark* racecar) and we converted and modified it to unmanned vehicles but retain the human driving function [1]. The car retains the basic characteristics of the formula car such as low center of mass. In addition, the vehicle has the high-voltage battery pack with 160kW peak power. Rear distributed motor drive system consists of the two high-power density wheel motor with peak 80Nm output torque. The two drive motors are independently controlled by respective motor controllers.

### A. Vehicle Modification

Several works need to be done to modify an electric formula car to an autonomous vehicle. The original vehicle is not driving by wire, team need to add steering and brake actuator. Since the rule says the autonomous formula car should keep fully functional and suitable for human driver, we reserve the mechanical steering system and add servo motor to drive the steering linkage. A servo brake system and a redundant emergency brake are equipped not only can perform the brake force control but also can stop the car safely when emergency or failure of the system happen. It is worth mentioning that a wireless emergency stop device is designed to control the brake and the tractive system power supply remotely and it is manipulated by the safety officer of the team. State indicator located at the top of the main loop shows the system state. Autonomous system switch can activate or stop the autonomous system and switch the vehicle mode between driver and driverless. Finally, all sensors and computation device are mounted according the rules. Fig. 2 shows the modifications on the vehicle.

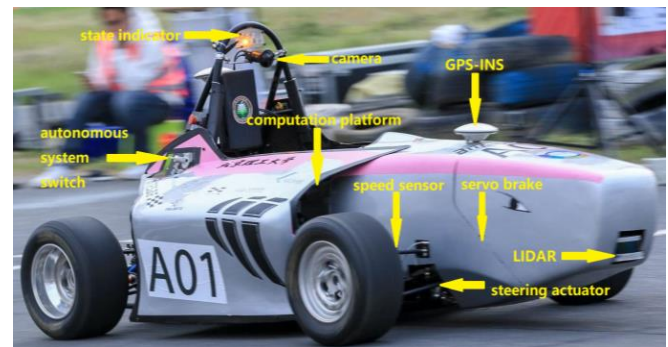

Fig. 2. Modifications including sensors, computer, actuators and indicator

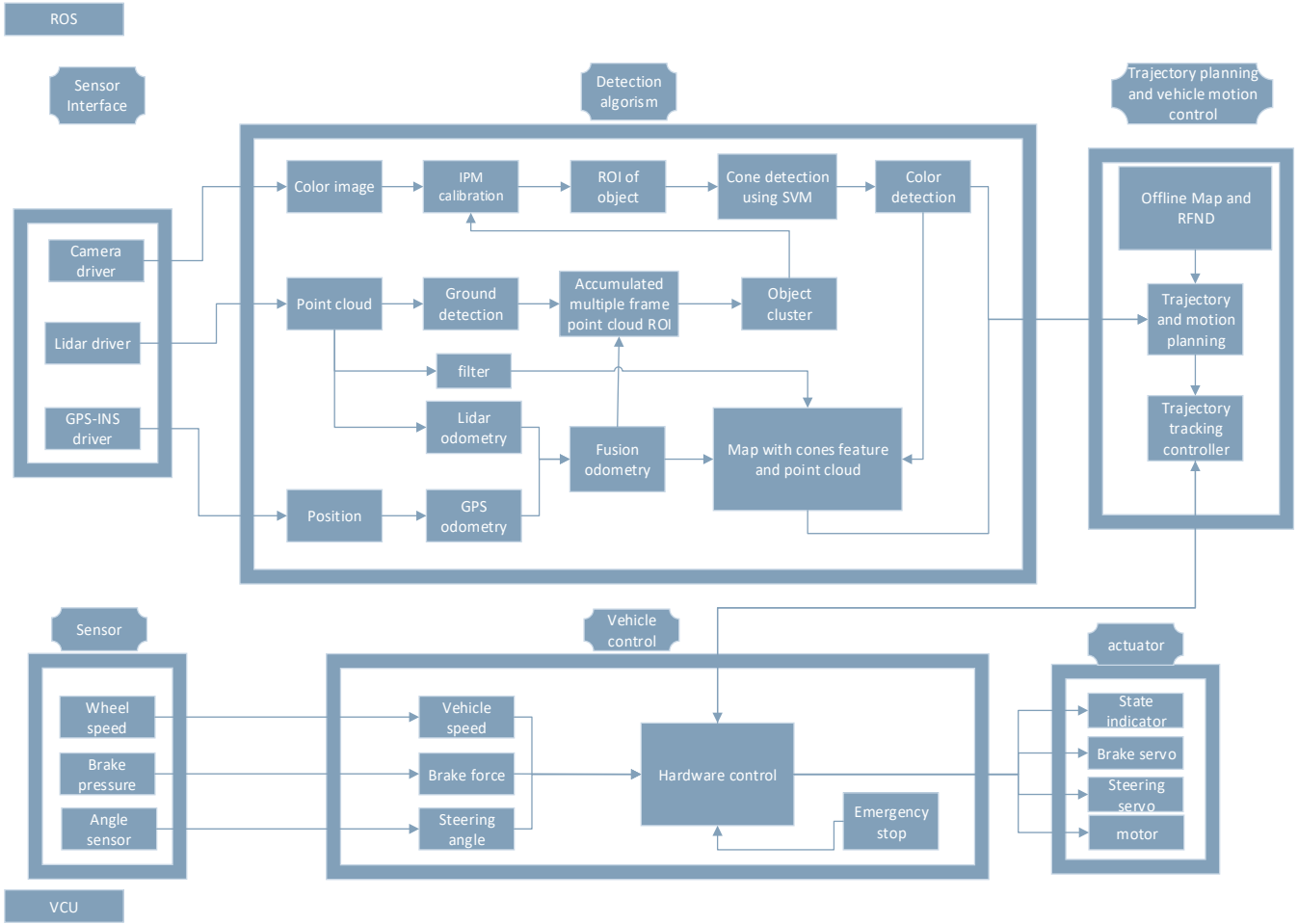

Fig. 3. Flow diagram of the software.

## B. Software Architecture

The software is basically divided in to two levels which have different task and run on different devices. The structure is shown in Fig. 3. The up-level runs on x86 platform with Linux OS and ROS software frame is installed to manage all the nodes and data communications. The up-level software includes sensor interface, detection node and planning and control node. Senor interface communicates with LIDAR, camera and GPS-INS and synchronize the data according to time label. The main function of detection node is point cloud processing for object detection, image processing of cones classification, odometry for vehicle motion estimation and mapping. The planning and control node is for motion control and sending control command such as steering angle and target speed to low-level controller via RS232. The low-level controller is for the hardware control task such as steering servo position control following the control command. In addition, the controller also controls the indictors, the tractive system and emergency stop according to the competition safety rules. We choose a rapid prototype VCU as the platform which has the resources to meet the requirement on computation and multi-interfaces. It has a 32-bit MCU with 264MHz router, multiple communication interface including CAN-BUS and RS232, PWM driven port and digital I/O. The VCU can be programmed by Simulink environment so the low level-development can be simplified.

## III. ENVIRONMENT PERCEPTION

In the FSAC, standard cones are the main track mark, which are 30cm high with three kinds of colors: red cones mark the left boundary of the cons and the blue cones mark the right while the yellow cones mark the start/finish points. So our detection algorithm is specialized to find cones and extract the color. Multiple sensors including LIDAR, camera and GPS-INS are cooperatively working in the mission.

### A. Laser Obstacle Detection

#### 1) Point cloud pre-processing

Choose ROI of the point cloud data. Since the mission mainly is detecting the object on the ground, we choose a box region with 1m height from ground and the size of 20mx20m.

In order to detect objects, the points reflect from ground plane must be detected and filtered. A simple algorithm for detecting obstacles and ground in laser scans would be to find points whose vertical displacement exceeds a given threshold. Indeed, this algorithm can be used to detect large obstacles such as pedestrians and cars. However, the lateral slop of the car body and calibration error are high enough that the displacement threshold cannot be set low enough in practice to detect cone-sized objects without substantial numbers of false positives from ground points [2].

To alleviate these problems, a ground detection method by RANSAC algorithm is implemented to find ground laser points and delete them from object ROI. We suppose that ground is a flat plane without large curve and bulge. Even though a plane does not present various terrains, it is sufficiently effective for our applications of a racing car. Most environments contain flat ground, and roads can be modeled as locally planar [3]. Moreover, the locally planar assumption is valid for our racing field. So a plane mode can be used as the mathematical model to describe ground plane. In this paper, we model ground as a 3D plane. Generally, a plane with three degree of freedom can be represented by four parameters as follows:

$$ax + by + cz + d = 0 \tag{1}$$

Beside the planar assumption, we also assume that the LIDAR placed in front of vehicles keeps almost constant height from ground. Only the small pitch and roll of ground plane are considered. The ground plane is written as:
$$y = kx + jz + h \tag{2}$$
Where h is approximately the height between the sensor and ground.

During the hypothesis evaluation, all the points in the ROI will be the input observed data. Error between a point and ground plane is commonly defined as their geometric distance,

$$\text{error}(p_i) = \frac{y - kx - jz - h}{\sqrt{1 + k^2 + j^2 + h^2}} \tag{3}$$

During each iteration, the label representing if a point is inlier point under the hypothesis plane parameters is defined as

$$l(p_i) = \begin{cases} 1 & if\ |error(p_i)| < \tau \\ 0 & else \end{cases} \tag{4}$$

After setting iterative number and fit threshold $\tau$, the algorithm can find the optimal estimation parameters of the plane and tell apart the inliers points whose distributions are fit in the solved plane mode and the noise data which are the laser points of the objects outside the ground plane. Fig. 4 shows the result. White points are in the ground plane while dark points represent the obstacles. It can be seen that the algorithm can distinguish ground and objects like cones.

*2) Accumulate multiple frame point cloud data*

After the pre-processing, the ROI of objects can be extracted. The next step could be the cluster and object extraction. However, the detection result will be limited in range and stability if only using single frame information. Firstly, LIDAR have sixteen layers and the vertical angle between layers is two degree. The low resolution will lead to a large non-detection zone between layers as the Fig. 5 shows. Secondly, the objects with small size or low height such as curbs and cones will be hard to detect and cluster due to few reflection points. Thirdly, the vertical vibration and ground fluctuations during driving lead to a high threshold to reduce

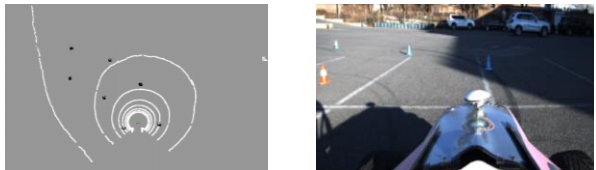
(a) (b)
Fig. 4. The white points are the inlier ground points and the dark points are the objects in (a) while (b) is the image from the camera.

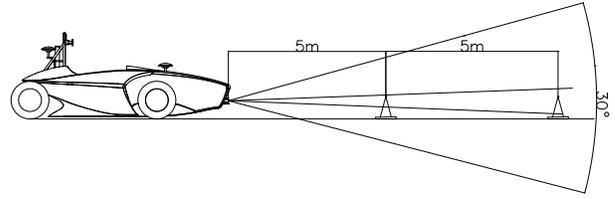
Fig. 5. The interval angle between tow scan layers can cause the dead zone. The laser cannot be sure to reach a cone located over 10m at any time.

the influence of the noise points.

In order to improve the effective recognition range of LIDAR and enhance the detection stability and reduce the influence of noise point, we use multi-frame accumulation algorithm which inspired by the strategy of building high definition maps [4]. After accumulation, the data points can become dense to improve the effective recognition range.

The coordinate transformation from previous frames coordinates to current frame coordinate need to be calculated which is the key part of the algorithm. The coordinate change is mainly caused by the three dimensional motion and 3 DOF pose transformation of the car since the LIDAR is mechanical fixed on the vehicle. In this paper, pose and translation matrix T are defined to describe the coordinate transformation.

Firstly, in order to calculate coordinate transformation matrix, the odometer information of the car including the position and heading vector in initial coordinate at the scan moment of each frame need to be calculated and matched as the position label.

Odometry contains the motion and position information including heading and coordinate in initial frame. Multiple kinds of sensors including monocular or binocular stereo vision sensor, LIDAR, GPS, INS and wheel speed sensor can be used as odometry. Wheel speed sensor is widely used as odometry on wheeled robot. However, the accuracy is limited by slip and wear of the tire. GPS can provide high accurate position in RTK mode under open out-door working condition. In recent years, many researches have been done on SLAM using vision system and LIDAR odometry as the front end [7]. In this paper, an odometry using LIDAR and GPS-INS coupled data is proposed. LIDAR and GPS-INS data are separately used to calculate the position change of the vehicle and coupled using complementary filter.

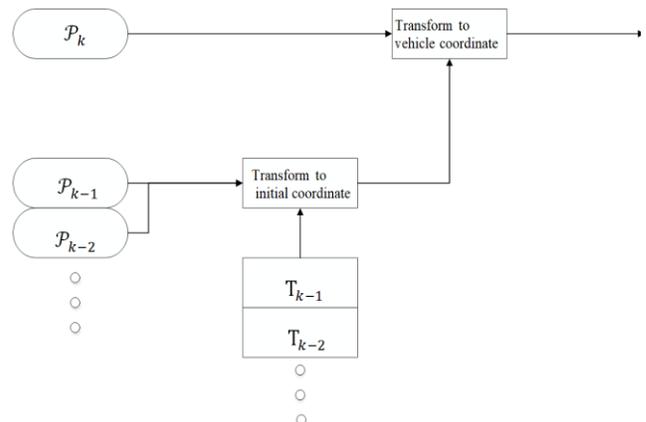
Fig. 6. The accumulation process flow chart. $\mathcal{P}_k$ is the point cloud acquired at $t_k$. $T_k$ is the pose-motion transformation to initial coordinate system.

The problem of LIDAR odometry is that, given the LIDAR clouds $\mathcal{P}_k$, k∈$Z^+$ arranged in chronological order, the registration of point cloud is calculated, and the movement of LIDAR in two scans is obtained. The whole algorithm is divided into three steps which inspired by LOAM, a high accurate LIDAR odometry [5.6]: The first step is to extract the registration feature points to improve the registration speed and accuracy. The algorithm selects the convex edge and the laser radar reflection point on the flat plane as the feature points. Set S is a set of continuous points i in the same scanning process, and define the evaluation formula of local surface smoothness as:

$$c = \frac{1}{|S| \cdot ||X_{k,i}||} \left\| \sum_{j \in S, j \neq i} (X_{k,i} - X_{k,j}) \right\| \quad (5)$$

Here, the maximum value greater than c indicates the edge point, and the minimum value less than c indicates the plane point. Taking c as the threshold to select feature points, the maximum and minimum thresholds can be obtained experimentally. The second step is the matching of feature points to complete point cloud registration. Considering that the distance between $t_{k+1}$ and the adjacent two times when $t_k$ moves is not too large, we need to find the nearest neighbor of each edge feature point or plane feature point in the point cloud set $\bar{\mathcal{P}}_k$ as Matching point. After obtaining the matching points of the feature points, we need to calculate the distance formula from the feature points to the matching points. We calculate the pose movement of the radar by minimizing the distances between all the feature points and the matching points. We use the ICP algorithm to solve this optimization problem. The positioning algorithm rated operating frequency of 10Hz.

Second step is to accumulate points to LIDAR coordinate system $\hat{L}$ at current time $\hat{t}$. After getting the odometry information, the coordinate transformation matrix to initial coordinate can be solved and the point cloud of the historical frame is transformed into the coordinate system of the current scan frame. Let the number of fusion frames be m. The point cloud set after multi-frame fusion is $\hat{\mathcal{P}}$ and the set of point cloud obtained at $t_k$ is $\mathcal{P}_k$, where k = 1, 2 ... m. The coordinate system of the point cloud set $\mathcal{P}_k$ is $L_k$ and the coordinate of the point i ∈ $\mathcal{P}_k$ under the $L_k$ is represented as $X_{i,k}$. According to the odometer information, we can get the pose-motion transformation $T_k$ from the initial system to the LIDAR coordinate system:

$$T_k = [t_x, t_y, t_z, x, y, z, \omega]^T \quad (6)$$

Where $t_x$, $t_y$ and $t_z$ are the translational changes in the x, y, and z axes and x, y, z, and ω are quaternions. For any point i ∈ $\mathcal{P}_k$, k = 1,2 ...... m, the fused point $\hat{X}$ cloud set $\hat{\mathcal{P}}$ can be obtained as:

$$\widehat{X_{i,k}} = (\hat{R} R_k^{-1} (X_{i,k} - T_k(1:3)) + \hat{T}(1:3)) \quad (7)$$

Where $R_k$ and $\hat{R}$ is the posture transformation matrix from initial coordinate system to LIDAR coordinate $L_k$ and $\hat{L}$, separately. When a Transformation T is given, R is defined as:

$$R = \begin{bmatrix} 1-2(y^2+z^2) & 2(xy-zw) & 2(xz+yw) \\ 2(xy+zw) & 1-2(x^2+z^2) & 2(yz-xw) \\ 2(xz-yw) & 2(yz+xw) & 1-2(x^2+y^2) \end{bmatrix} \quad (8)$$

*3) Cluster point cloud*

After accumulating the points of objects, Euclidean cluster extraction method is used to divide point cloud reflected from the same object into groups and compute the centroid location. What's more, the reflection numbered points and the envelope size of the object can be calculated after cluster so that we can filtrate the object groups with excessively large size compared to traffic cones.

After the cluster and filtration by size, the cones can be detected. However, there are still other objects like tyre stacks around the track which can be detected as cones. So in the next section, the vision information is used to verify.

*B. Vision Based Obstacle classification*

In our work, in order to improve the real time computation, a LIDAR-assistant visual detection algorithm is implemented. Following three steps show the main process:

*1) Joint calibration of LIDAR and camera*

To achieve LIDAR-assistant detection, the picture is projected onto a new viewing plane by perspective transformation to project object position in image space onto LIDAR coordinate. We can construct the perspective matrix between the two planes by choosing four pairs of different corresponding points:

$$[x' \quad y' \quad w'] = [u \quad v \quad w] \begin{bmatrix} T_1 & T_2 \\ T_3 & a_{33} \end{bmatrix} \quad (9)$$

Where: u and v represent the pixel coordinate of the bottom center of the cones. $x'/w'$ and $y'/w'$ represent the position of the cones in XoY plane. $T_1$ is the linear transformation. $T_2$ is Perspective effect segment and $T_3$ is translation matrix. Fig. 7(a) shows the aerial view after transformation. The bottom center points of the cones in mage coincident with the cluster result.

Through this matrix, it is possible to map all the cone-like objects in the real-world coordinates onto the camera plane and delineate the ROI of the corresponding obstacles on the image plane according to the mapped points, and the subsequent processing only processes these ROI regions Image. Fig. 7 shows the transformation process and the corresponding ROI bounding box.

*2) Cones detection based on SVM*

The HOG features are calculated for each gray scale converted by the ROI. HOG feature is to select the pixel block, and then calculate the gradient value in each direction of the pixel block, and finally generate the corresponding length of the feature vector from these gradient values, which can be

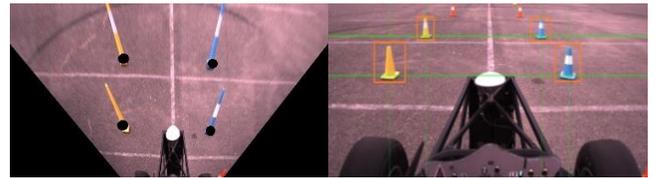

(a) Aerial View　　　　　　(b) ROI selection
Fig. 7. Joint calibration by perspective transformation and ROI selection

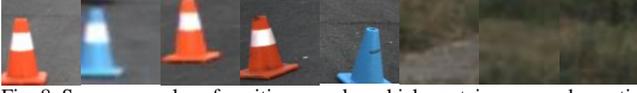
Fig. 8. Some examples of positive samples which contain cone and negative samples such as grass or curbs from training dataset.

used as the classification basis to train the SVM classifier. We build the dataset of the cones and train the model using OpenCV tools. Fig. 8 shows the examples from dataset.

*3) Cone color detection*

In order to obtain the cones color information in the ROI, it is necessary to first determine whether the boxed cone analogue matches the color feature and the color of the cone inside the ROI. First, the influence of light on the camera imaging needs to be reduced. In the HSV color space, H represents the hue, S represents the saturation, V represents the brightness of the color, and the following conversion method can be used to convert from the RGB color space to the HSV color space:

$$\begin{aligned} R' &= R/255 \\ G' &= G/255 \\ B' &= B/255 \end{aligned} \quad (10)$$

$$\begin{aligned} C_{max} &= \max(R', G', B') \\ C_{min} &= \min(R', G', B') \\ \delta &= C_{max} - C_{min} \end{aligned} \quad (11)$$

H, S and V channels can be calculate as:

$$H = \begin{cases} 0°, & \delta = 0 \\ 60° \times (\frac{G'-B'}{\delta} \bmod 6), & C_{max} = R' \\ 60° \times (\frac{B'-R'}{\delta} + 2), & C_{max} = G' \\ 60° \times (\frac{R'-G'}{\delta} + 4), & C_{max} = B' \end{cases} \quad (12)$$

$$S = \begin{cases} 0, & C_{max} = 0 \\ \frac{\delta}{C_{max}}, & C_{max} \neq 0 \end{cases} \quad (13)$$

$$V = C_{max} \quad (14)$$

It can be seen that if do not use the V component, it can greatly reduce the impact of light on cone color recognition. After using the k-means clustering algorithm and taking k as 2, all the pixels are grouped into two categories. The cones color can be obtained by extracting the main color components in the ROI.

## IV. EXPERIMENT

The data is recorded during one track-drive test run. In this item, the track will be a loop with 4m width and 5m distance between one pair of cones as Fig. 9 shows. During each ride, the vehicle need finish two entire loops without any prior data

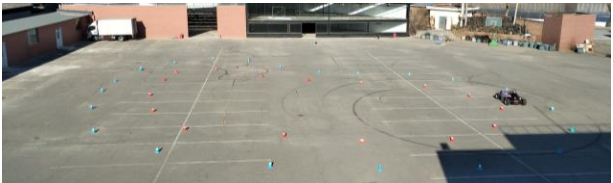
Fig. 9. The test field environment and track

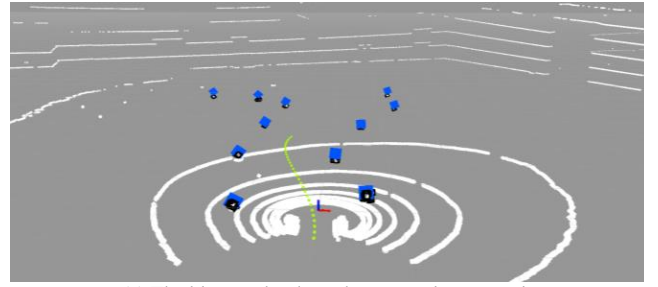
(a) The blue marks show the cones cluster result

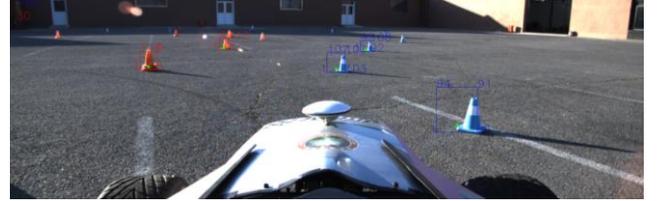
(b)The SVM classification and color detection result
Fig. 10. The cones detection result

of the track and stop after the finish line automatically. The fastest lap time will be recorded.

In the first lap, the vehicle will operate using LIDAR and visual system as real-time cones perception. The motion control will be based on the real-time planning algorithm to follow the middle line of the track as Fig. 10 shows. In the test, the odometry and mapping node is also run simultaneously to record the trajectory and build the map of the track as Fig. 11 shows. When the program detects a close loop of the trajectory, the vehicle will automatically shift its control mode.

In the second lap, the vehicle will mainly use GPS-INS to positioning and tracking the trajectory generated in first lap and the real-time planning just running as backup and will take over under receiving bad GPS data. Since the whole loop trajectory is known, the tracking algorithm can planning the speed precisely at each waypoint. What's more, the vehicle can run steadily without the effect of the detection error. So second lap time will have a greatly improved and the vehicle can reach a higher speed. Besides, the GPS-INS has higher output rate and lower delay than other perception system which help to perform a precise and robust control. Fig. 12 to Fig. 16 shows the data of the second lap.

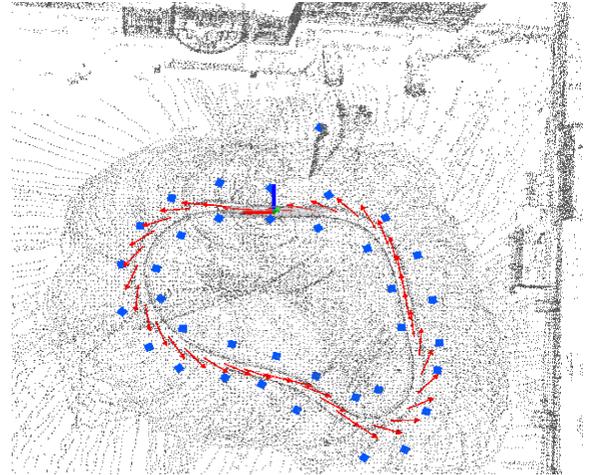
Fig. 11. After finish the first lap, the map of the track is built. Blue boxes show the position of the cones. The odometry data of each moment is represented by red arrow.

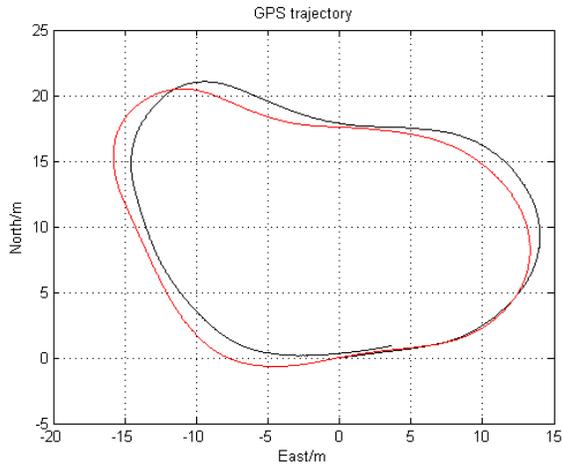

Fig.12. The black trajectory shows the ground truth of the target trajectory. The red trajectory shows the tracking result in the second lap.

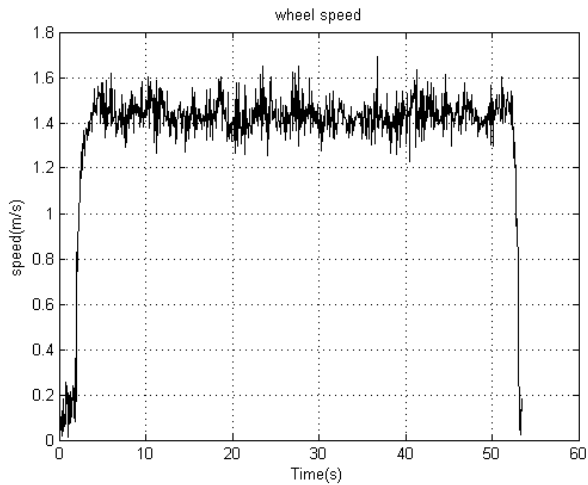

Fig 13. The vehicle speed from wheel speed sensor. The vehicle keeps a constant speed as the program setting.

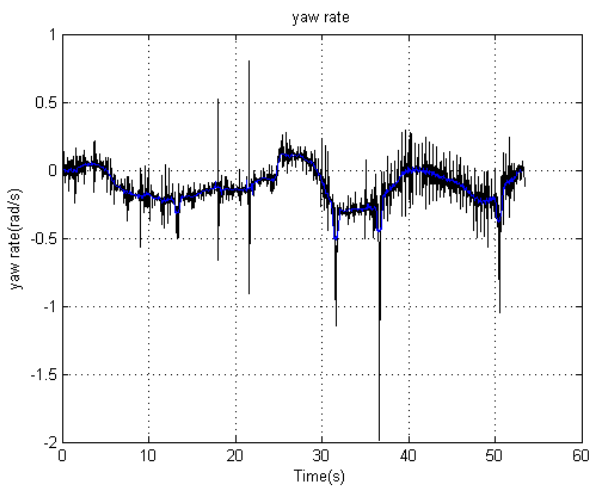

Fig. 14. The yaw rate data from GPS-INS.

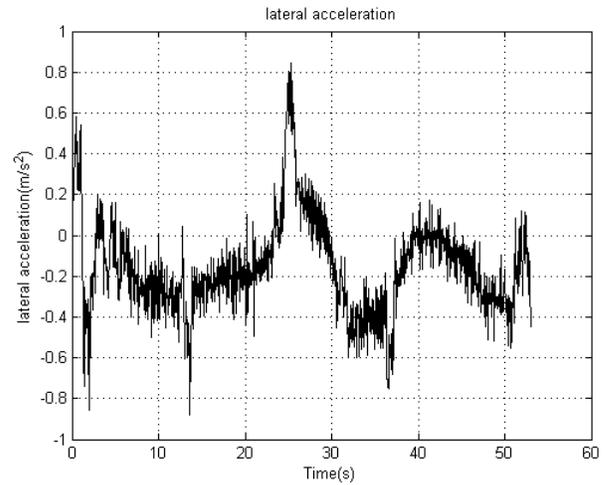

Fig. 15. The lateral acceleration data from GPS-INS.

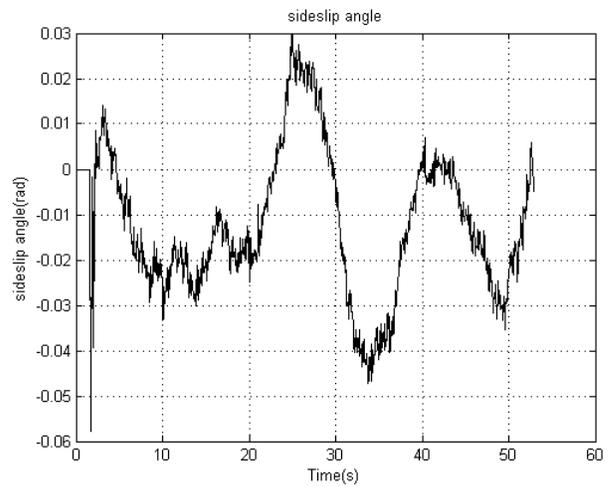

Fig. 16. The sideslip angle of the vehicle

## REFERENCES

[1] Ni, Jun, and J. Hu. "Path following control for autonomous formula racecar: Autonomous formula student competition." Intelligent Vehicles Symposium IEEE, 2017:1835-1840.
[2] Montemerlo, Michael, et al. "Junior: The Stanford entry in the Urban Challenge." Journal of Field Robotics 25.9(2009):569-597.
[3] Choi, Sunglok, et al. "Robust ground plane detection from 3D point clouds." International Conference on Control, Automation and Systems IEEE, 2014:1076-1081.
[4] Aldibaja, Mohammad, N. Suganuma, and K. Yoneda. "LIDAR-data accumulation strategy to generate high definition maps for autonomous vehicles." IEEE International Conference on Multisensor Fusion and Integration for Intelligent Systems IEEE, 2017:422-428.
[5] Zhang, J., and S. Singh. "LOAM : Lidar Odometry and Mapping in real-time." (2014).
[6] Zhang, Ji, and S. Singh. "Low-drift and real-time lidar odometry and mapping." Autonomous Robots 41.2(2016):1-16.
[7] Zhang, Ji, and S. Singh. "Visual-lidar odometry and mapping: low-drift, robust, and fast." IEEE International Conference on Robotics and Automation IEEE, 2015:2174-2181.